\documentclass{article}

\usepackage{microtype}
\usepackage{graphicx}
\usepackage{subfigure}
\usepackage{booktabs} % for professional tables
\usepackage{amssymb}
\usepackage{amsmath}
\usepackage{booktabs}
\usepackage{multirow}

\usepackage{hyperref}

\usepackage[accepted]{icml2020}

\icmltitlerunning{MxPool: Multiplex Pooling for Hierarchical Graph Representation Learning}

\begin{document}

\twocolumn[
\icmltitle{MxPool: Multiplex Pooling for Hierarchical Graph Representation Learning}




\begin{icmlauthorlist}
\icmlauthor{Yanyan Liang}{ed}
\icmlauthor{Yanfeng Zhang}{ed}
\icmlauthor{Dechao Gao}{ed}
\icmlauthor{Qian Xu}{ed}
\end{icmlauthorlist}

\icmlaffiliation{ed}{Northeastern University, China}

\icmlcorrespondingauthor{Yanyan Liang}{1801792@stu.neu.edu.cn}
\icmlcorrespondingauthor{Yanfeng Zhang}{zhangyf@mail.neu.edu.cn}

\icmlkeywords{Machine Learning, ICML}

\vskip 0.3in
]



\printAffiliationsAndNotice{}  % leave blank if no need to mention equal contribution

\begin{abstract}
How to utilize deep learning methods for graph classification tasks has attracted considerable research attention in the past few years. Regarding graph classification tasks, the graphs to be classified may have various graph sizes (i.e., different number of nodes and edges) and have various graph properties (e.g., average node degree, diameter, and clustering coefficient). The diverse property of graphs has imposed significant challenges on existing graph learning techniques since diverse graphs have different best-fit hyperparameters. It is difficult to learn graph features from a set of diverse graphs by a unified graph neural network. This motivates us to use a multiplex structure  in a diverse way and utilize a priori properties of graphs to guide the learning. In this paper, we propose MxPool, which concurrently uses multiple graph convolution/pooling networks to build a hierarchical learning structure for graph representation learning tasks. Our experiments on numerous graph classification benchmarks show that our MxPool has superiority over other state-of-the-art graph representation learning methods.
\end{abstract}

\newcommand{\Paragraph} [1] {\smallskip\noindent{\bf #1. }}

\section{Introduction}

%Graph classification is an important problem with practical applications in a diverse set of fields. The task is to identifying the class labels of graphs in a dataset.

Graphs are known to have complicated structures and have myriad of real world applications. Recently, great efforts have been put on utilizing deep learning methods for graph data analysis. Many newly proposed graph learning approaches are inspired by Convolutional Neural Networks (CNNs) \citep{LeCun:1998:CNI:303568.303704}, which have been greatly successful in learning two-dimensional image data (grid structure). The convolution and pooling layers in CNNs have been redefined to process graph data. Multitude of different Graph Convolutional Networks (GCNs) \citep{Shuman2013} have been proposed, which can learn node level representations by aggregating feature information from neighbors (spatial-based approaches) \citep{graphsage2017} or by introducing filters from the perspective of graph signal processing (spectral-based approaches) \citep{gcnspectral}. On the other hand, similar to the original pooling layer which comes with CNNs, graph pooling module \citep{chebnet2016,sortpool2018} could easily reduce the variance and computation complexity by downsampling from original feature data, which is of vital importance, particularly for graph level classification tasks. Recently, hierarchical pooling methods that can learn hierarchical representations of graphs have been proposed \citep{diffpool,graphunet,sagpool} and shows state-of-the-art performance for graph classification tasks.

%These methods generate deep GNNs by ``stacking'' GNN layers in a hierarchical fashion, each layer coarsening the input graph more and more.

 %\cite{gnn2009,DBLP:journals/corr/abs-1901-00596,DBLP:journals/corr/abs-1812-08434,DBLP:journals/corr/abs-1812-04202,DBLP:journals/corr/abs-1806-01261}.

%A common solution for augmenting the CNN convolution layers would be to use multiple convolution kernels in order to learn multiple local features. By contrast, most CNN pooling schemes are based on heuristics (e.g. a single max pooling kernel or sum pooling kernel) and have no clear link to the cost function of the model. While for graph pooling schemes, DiffPool \cite{diffpool} learns how to cluster together nodes to build a hierarchical multi-layer scaffold on top of the underlying graph, which learns a differentiable soft node-cluster assignment at each layer. Compared to standard CNNs, the differentiability of pooling operation is especially necessary for graphs due to the complex topological structure of graphs.

%Graphs contain no natural notion of spatial locality, i.e., one cannot simply pool together all nodes in a ``$m\times m$ patch'' on a graph.

However, the diverse property of graphs have imposed significant challenges on existing graph representation learning techniques. The graphs to be learned have various graph sizes (i.e., different number of nodes and edges) and have various graph properties (e.g., average node degree, diameter, clustering coefficient, etc.). Using unified graph neural network with consistent hyperparameters can bring troubles in the graph convolution operation as well as the graph pooling operation.

\begin{figure}[t]
	\centerline{
	\subfigure[Small graph]{\includegraphics[width=1.4in]{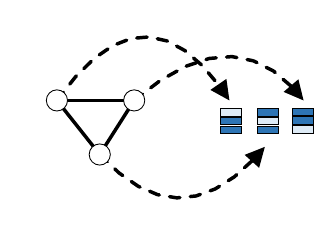}
    \label{fig:dimsmall}
    \vspace{-0.05in}}
    \hspace{2mm}
    \subfigure[Large graph]{\includegraphics[width=1.6in]{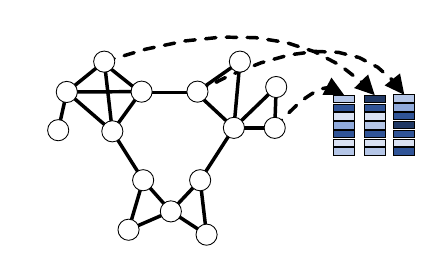}
    \label{fig:dimlarge}
    \vspace{-0.05in}}
    }
    \vspace{-0.1in}
	\caption{Multiple graph convolution networks with different hidden/output node embedding sizes are needed for graphs with various sizes. Small graph requires small node embedding size, while large graph requires large node embedding size.}
	\label{fig:dim}
\vspace{-0.1in}
\end{figure}

For example, when performing node-representation learning tasks (by graph convolution operation), it is enough to use small output embedding size for simple and small graphs, as shown in Figure \ref{fig:dimsmall}, since large embedding size could result in overfitting problem. By contrast, it is necessary to set large output embedding sizes for complex and large graphs to learn complex graph structure properties, as shown Figure in \ref{fig:dimlarge}. This will create a contradiction when processing a set of irregular heterogeneous graphs. For another example, when coarsening graphs (by graph pooling operation) for graph-representation learning tasks, different-size graphs are coarsened to consistent-size graphs in order to finally obtain fixed-size embedding vectors. According to a state-of-the-art graph pooling model DiffPool \cite{diffpool}, the graph coarsening process is based on two factors, the largest graph size $x$ among all graphs (i.e., the maximum number of nodes) and the coarsening ratio $r$, and all graphs are coarsened to graphs with $x\cdot r$ number of nodes. This will lead to unexpected node split operations (upsampling) instead of node merge operations (downsampling) and bring inaccuracies during the pooling process. But if the coarsening ratio $r$ is set too small, though the upsampling for small graphs can be avoided, the large graphs will lose too much information during the hierarchical coarsening process. Thus, this will also bring troubles when processing a set of irregular heterogeneous graphs.

\begin{figure}[t]
%\vspace{-0.1in}
	\centerline{
	\subfigure[Small graph]{\includegraphics[width=1.5in]{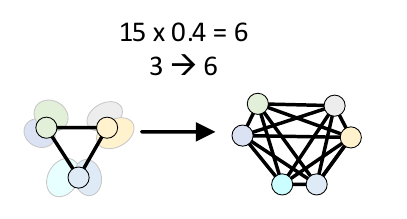}
    \label{fig:poolsmall}
    \vspace{-0.05in}}
    \hspace{-0.2in}
    \subfigure[Large graph]{\includegraphics[width=2in]{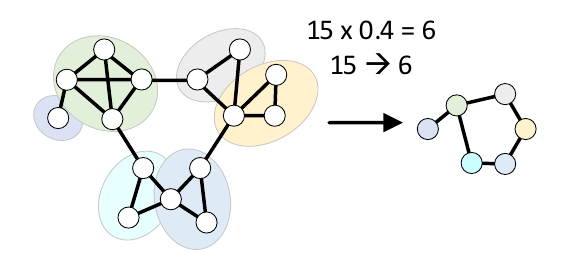}
    \label{fig:poollarge}
    \vspace{-0.05in}}
    }
    \vspace{-0.1in}
	\caption{Multiple graph pooling networks with different coarsening ratios are needed for graphs with various sizes. Suppose the largest graph size is 15 and the coarsening ratio is 0.4. After pooling, all the coarsened graphs should have $15\times 0.4=6$ nodes. It is good for large graphs, but it will bring troubles for small graphs. For example in (a), the 3 nodes in a small graph are split in order to have 6 nodes, which will incur inaccuracies during the ``coarsening'' process. Thus, small graph requires small coarsening ratio (e.g., 0.2), while large graph requires large coarsening ratio.}
	\label{fig:pool}
\vspace{-0.1in}
\end{figure}

%\begin{figure*}[t]
%\label{fig:motivation}
%  \centering
 % \includegraphics[width=5.5in]{pic/motivation.eps}\\
 % \caption{Multiple graph convolution networks with different output node embedding sizes are needed since graph sizes and complexities are not consistent, e.g., (a) and (b). Multiple graph pooling networks are needed to coarsen graphs according to different graph properties. For example, the original graph can be coarsened according to graph structure as shown in (c) or according to node features as shown in (d), where circle vertices have similar node features and triangle vertices have similar node features.}
%\end{figure*}

\begin{table}[t]
	\caption{Effect of embedding size (dim) on various-size graphs (DiffPool \cite{diffpool} on PROTEINS dataset).}
	\vspace{0.05in}
	\label{tab:dim}
	\centering
	{\begin{tabular}{l c c c}	
    \toprule[1pt]
		{graph size} &
		{dim=10} &
		{dim=60} &
        {dim=120}\\
		\toprule[0.5pt]
		{small} & \textbf{74.70} & 73.91 & 73.55 \\
		{medium} & 71.56 & \textbf{73.72} & 71.47 \\
		{large} & 81.32 & 81.57 & \textbf{82.72} \\	
    \toprule[1pt]
	\end{tabular}
	}
	\vspace{-0.1in}
\end{table}

\begin{table}[t]
    \vspace{-0.1in}
	\caption{Effect of coarsening ratio $r$ on various-size graphs (DiffPool \cite{diffpool} on PROTEINS dataset).}
	\vspace{0.05in}
	\label{tab:pool}
	\centering
	{\begin{tabular}{l c c c}	
    \toprule[1pt]
		{graph size} &
		{$r=0.05$} &
		{$r=0.1$} &
        {$r=0.15$}\\
		\toprule[0.5pt]
		{small} & \textbf{76.34} & 71.76 & 73.91 \\
		{medium} & 73.23 & \textbf{74.41} & 73.50 \\
		{large} & 82.63 & 82.11 & \textbf{83.42} \\	
    \toprule[1pt]
	\end{tabular}
	}
	\vspace{-0.15in}
\end{table}

To verify our analysis, we divide the PROTEINS graphs dataset \citep{proteingraph} into three subsets of graphs according to their sizes (number of nodes) and run a state-of-the-art graph pooling model DiffPool \cite{diffpool} on the three subsets of graphs respectively to perform graph classification task. The PROTEINS dataset is evenly partitioned into three subsets, where the graphs with $[0,20]$ nodes are considered as small graphs, the graphs with $[21,35]$ nodes are considered as medium graphs, and the graphs with $[36,200]$ nodes are considered as large graphs. The average accuracy results\footnote{The result for each dimension is obtained by averaging multiple results with different coarsening ratios $r=0.05,0.1,0.15$.} with different embedding size parameters ($dim=10,60,120$) on the three subsets are listed in Table \ref{tab:dim}. We can see that small graphs prefer small embedding size while large graphs prefer large embedding size as expected. Furthermore, we obtain better results on large graphs. We then choose the best embedding size parameter for different size graphs (dim=10 for small graphs, dim=60 for medium graphs, and dim=120 for large graphs) and vary the coarsening ratio settings to see the accuracy results on different size graphs. The results with different coarsening ratio parameters ($r=0.05,0.1,0.15$) are listed in Table \ref{tab:pool}. As expected, the small graphs prefer small coarsening ratio, while the large graphs prefer large coarsening ratio.

The diverse property of graphs dataset and its effect to the preference on various GNN hyperparameter settings motivate us to use multiplex GNN structure to learn graph features in a diverse way. As known, a common solution for augmenting the traditional CNN convolution layers is to use multiple convolution kernels in order to learn multiple local features. The success of CNNs on image data also inspires us to concurrently use multiple graph convolution networks and multiple graph pooling networks to learn graph representations. Besides, the effect of graph size on the performance also motivate us to utilize a priori properties of graphs to guide the learning process.

%As the differentiability of pooling is necessary for graphs and the multi-kernel convolution in CNN have shown improved performance, one question is accordingly raised: Why not using multiple convolution kernels and multiple differential pooling kernels to improve representation learning for the more complex graph structures?

In this paper, we propose MxPool in hierarchical graph representation learning for graph classification tasks\footnote{Our code is available at https://github.com/JucatL/MxPool/.}. MxPool comprises multiple graph convolution networks (with different hyperparameters) to learn node-level representations and also comprises multiple graph pooling networks to coarsen the graph. The node-level representations resulted from multiple convolution networks and the coarsened graphs resulted from multiple pooling networks are merged in a learnable way, respectively. The a priori properties of graphs are considered during the merge process, where the to-be-merged vectors/matrices from different GNNs are endowed with different weights. The merged node representations and the merged coarsened graph are then used to generate a new coarsened graph, which is used in the next layer. This multiplex structure can adapt to graphs with diverse properties and can extract useful information from different perspectives.

%\footnote{Our code is available at https://github.com/JucatL/MxPool/.}

%The multi-kernel pooling comprises of pooling kernels with multiple receptive fields to capture different soft cluster assignments, then, weightedly stacking these assignment matrices together to its successive layer. Introducing multiple pooling kernels to hierarchical graph representation learning is important due to the following reasons. 1) Since graphs are more complex with not only node/edge features but also topological features, multi-kernel pooling can learn multiple node-cluster assignments by considering different feature combinations (i.e., node features and graph structures); 2) Since graph sizes are not consistent (with varying numbers of nodes and edges), multi-kernel pooling can capture multi-scale graph features.

We conduct extensive experiments on numerous graph classification benchmarks and show that our MxPool has marked superiority over other state-of-the-art graph representation learning methods. For example, MxPool achieves 92.1\% accuracy on the D\&D dataset while the second best method DiffPool only achieves 80.64\% accuracy.

%We summarize Our results on graph classification demonstrate that MKPool can achieve superior performance over the state-of-the-art methods GraphSage \cite{graphsage2017}, GCN, SortPool \cite{sortpool2018}, DiffPool \cite{diffpool}, SAGPool \cite{sagpool}, and gPool \cite{graphunet}.

%We introduce a novel pooling operator Multi-Kernel Pooling (MKPool), which can extracts graph features from different angles;
%We incorporate pooling layers based on MKPool into existing graph neural networks as a novel framework MKGCN for representation learning for graph classification;

\section{Related Work}

In this section, we review the recent literature on GNNs, graph convolution variants, and graph pooling variants.

\Paragraph{Graph Neural Networks Inspired by Traditional Deep Learning Techniques}GNNs have recently drawn considerable attention due to their superiority in a wide variety of graph related tasks, including node classification \citep{nodeclassification}, link prediction \citep{linkprediction}, and graph classification \citep{Dai:2016:DEL:3045390.3045675}. Many of these GNN models are inspired by traditional learning techniques. Inspired by the huge success of convolutional networks in the computer vision domain, a large number of Graph Convolutional Networks (GCNs) have emerged. Besides convolution operation, pooling operation, as another key component in CNNs, has also inspired research communities to propose graph pooling operations. There are also GNN optimizations originating from other learning approaches. Inspired by Recurrent Neural Networks (RNNs), \citet{graphrnn2018} apply Graph RNN to the graph generation problem. DGNN \citep{DGNN} proposes using LSTM \citep{lstm1997} to learn node representations in dynamic graphs. Inspired by the attention mechanism \citep{attention2017} Graph Attention Networks (GATs) \citep{gat2017} introduce attentions into GCNs by differentiating the influence of neighbors. Graph AutoEncoders (GAEs) \citep{SDNE2016} origin from the autoencoder mechanism widely used for unsupervised learning and are suitable to learn node representations for graphs. GCPN \citep{GCPN2018} utilizes Reinforcement Learning (RL) for goal-directed molecular graph generation.

\Paragraph{Graph Convolution}Graph convolution operations fall into two categories, spectral-based approaches and spatial-based approaches. \citet{gcnspectral} first introduce convolution for graph data from spectral domain using the graph Laplacian matrix \textbf{$L$}. Besides, there exist numerous spectral-based graph convolution methods, such as ChebNet \citep{chebnet2016}, 1stChebNet \citep{1stchebnet}, and AGCN \citep{AGCN2018}. In contrast, spatial-based convolution methods define graph convolution based on a node's spatial relations. It takes the aggregation of a node representation and its neighbors' representations to obtain a new representation for this node. In order to explore the depth and breadth of a node's receptive field, multiple graph convolution layer are stacked together, so that the features of two or more hops away neighbors can be learned. For example, GGNNs \citep{ggcn}, MPNN \citep{mpnn2017}, GraphSage \citep{graphsage2017}, PATCHY-SAN \citep{PATCHY-SAN2016}, and DCNN \citep{DCNN2016} all fall into the spatial-based category.

\Paragraph{Graph Pooling}Graph pooling operation is of vital importance for graph classification tasks \citep{sortpool2018}. It coarsens a graph into sub-graphs \citep{chebnet2016,diffpool} or to sum/average over the node representations \citep{Duvenaud:2015:CNG:2969442.2969488,mpnn2017}, which can obtain a compact representation on graph level. The graph coarsening approaches obtain hierarchical graph representations either by using \textit{deterministic} pooling methods or by using \textit{learned} pooling methods. The deterministic pooling methods \citep{chebnet2016,DBLP:journals/corr/SimonovskyK17} utilizes graph clustering algorithms to obtain next level coarsened graph that is going to be processed by GNNs, following a two-stage approach. On the other hand, the learned pooling methods \citep{diffpool,sagpool,edgepool,graphunet} seek to learn the hierarchical structure, which have shown to outperform deterministic pooling methods. DiffPool \citep{diffpool} was the first to propose learned graph pooling. It learns a soft cluster assignment matrix in layer $l$ which contains the probability values of nodes being assigned to clusters. A cluster in layer $l$ will be reduced to a node in layer $l+1$. A GNN with input node features and adjacency matrix is used to generate the soft assignment matrix, based on which we can learn the cluster embeddings (i.e., node features in the next layer) and the coarsened adjacency matrix denoting the connectivity strength between each pair of the clusters. Besides DiffPool, numerous graph pooling methods have emerged recently, including gPool \citep{graphunet}, SAGPool \citep{sagpool}, EigenPooling \citep{Ma:2019:GCN:3292500.3330982}, Relational Pooling \citep{relatepool}, and StructPool \citep{structpool}. However, to the best of our knowledge, none of the existing pooling methods employs multiplex structure to learn graph representations in a diverse way.

\section{Proposed Method}
\label{sec:method}

In this section, we propose MxPool to learn graph representations for graph level classification tasks. Before going to the details, we first introduce some notations and the problem setting.

\begin{figure*}[t]
%\vspace{-0.1in}
\label{fig:overall}
  \centering
%  % Requires \usepackage{graphicx}
  \includegraphics[width=6.7in]{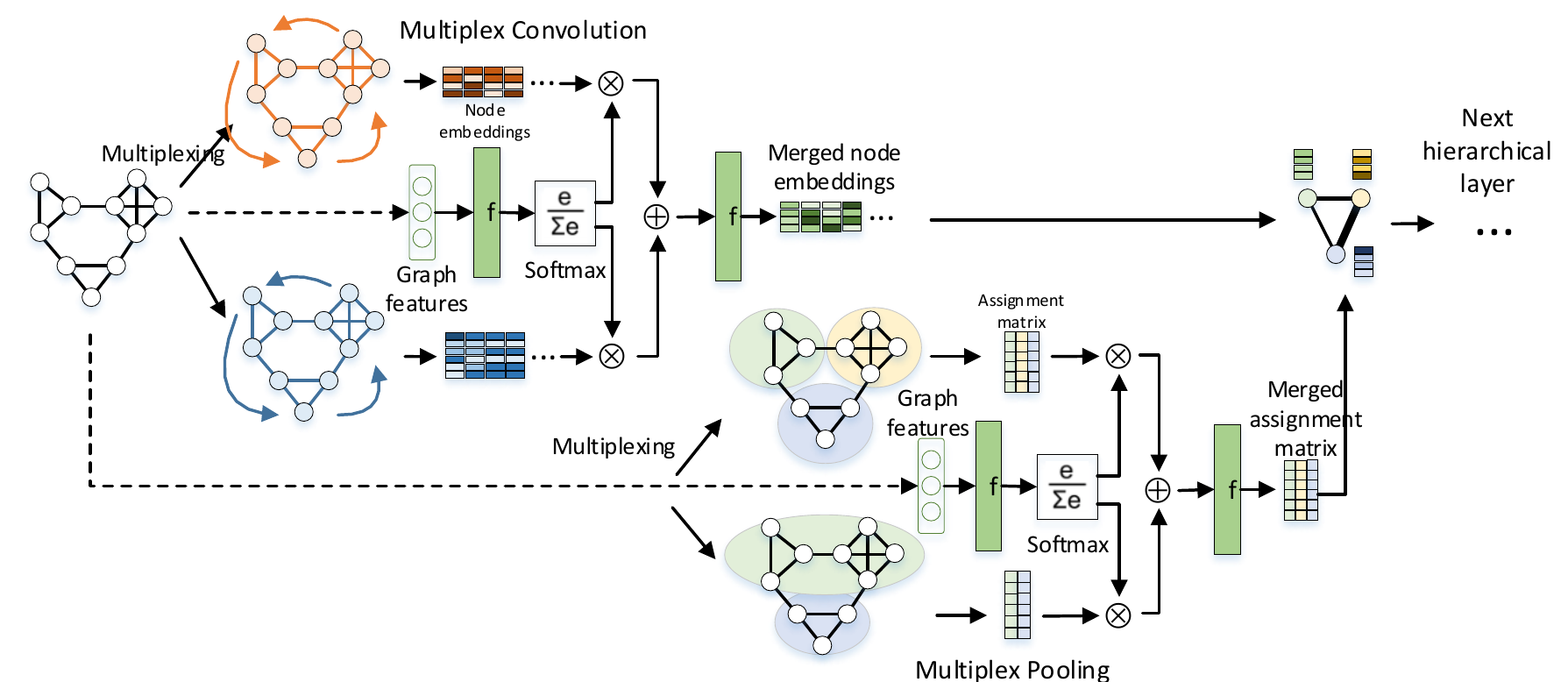}\\
  %\vspace{-0.1in}
  \caption{An illustrative example of MxPool. At each hierarchical layer, we first run multiple GCNs with different hyperparameters on the original graph, so we will have multiple diverse sets of node embeddings. The a priori graph properties are used to guide the merge of different sets of node embeddings. The better-fit GCN's results are given higher weight (by `$\otimes$'), where the weights are learned by a softmax layer. The node embeddings from different GCNs are concatenated (`$\oplus$') as the input of a fully connected layer, which will output the merged node embeddings. We also run multiple differentiable pooling operations on the original graph, and cluster nodes together to obtain multiple node-cluster assignment matrices. These assignment matrices are also merged into one with the guidance of the a priori graph properties in a learnable way. The merged node embeddings and the merged assignment matrix are used to generate a new coarsened graph with a new set of node embeddings, which is going to be processed in the next hierarchical layer.}
  \vspace{-0.1in}
\end{figure*}

\newcommand{\Paragraphnopunc}[1]{\smallskip\noindent{\bf #1}}

\Paragraphnopunc{Problem Setting} A graph can be represented as $\mathcal{G}=\{A,F\}$, where $A\in\mathbb{R}^{n\times n}$ denotes the adjacency matrix ($n$ is the number of nodes contained in $\mathcal{G}$), and $F\in\mathbb{R}^{N\times d}$ denotes the node feature matrix ($d$ is the dimension of features). In the graph classification setting, given a set of graphs and each being associated with a label, we aim to train a model that takes an unseen graph as input and predicts its corresponding label. To make the prediction, it is important to extract useful information from multiple perspectives including both graph structure and node features.

\subsection{Overview}

MxPool is a multi-layer hierarchical GNN model. At each layer, MxPool consists of convolution operation and pooling operation. The convolution operation aims to learn node-level representations, while the pooling operation aims to learn a coarsened graph. The new coarsened graph can then be used as input to next layer. This process can be repeated several times, generating a multi-layer GNN model to learn hierarchical graph representations. The convolution operation and pooling operation are both important for graph representation learning. To simplify the illustration, we choose GCN \citep{nodeclassification} as the convolution layer and DiffPool \citep{diffpool} as the pooling layer (which is a differentiable pooling method), but it can be extended to use other convolution/pooling variant as well.

Different from other hierarchical GNNs, MxPool launches multiple GCNs to learn node-level representations and also launches multiple pooling networks to coarsen the graph. The node-level representations resulted from multiple GCNs are then merged in a learnable way by considering the a priori graph properties (e.g., number of nodes/edges, diameter, average node degree, etc.), and the coarsened graphs resulted from multiple pooling networks are also merged. The merged node embeddings and the merged coarsened graph are used to generate a new coarsened graph with a new set of node features. This multiplex structure can help extract useful information from different perspectives and can adapt to graphs with different sizes. We provide an illustrative example as shown in Figure \ref{fig:overall}.

The procedure of the GCN \citep{nodeclassification} is to ``horizontally'' learn node representations, as it can only ``pass message'' between nodes through edges. The procedure of differentiable pooling \citep{diffpool} is to ``vertically'' summarize the node features into the higher level graph representation. The procedure of multiplexing is to ``diversely'' learn node representations or graph representations from different perspectives. The procedure of merging is to ``synthetically'' learn the diverse results and put more attention to one or more perspectives according to the a priori graph features. Since the convolution, pooling, and merging operations are all differentiable, we can define an end-to-end differentiable graph representation learning framework in a hierarchical manner.

\subsection{Multiplex Convolution}

In our model, we use GCN for the convolution operation. The original GCN \citep{nodeclassification} is stacked by several convolutional layers, and a single convolutional layer can be written as
\begin{equation}\label{eq:gcn}
  H^{(k+1)}=ReLU(\tilde{D}^{-\frac{1}{2}}\tilde{A}\tilde{D}^{-\frac{1}{2}}H^{(k)}W^{(k)}),
\end{equation}
$H^{(k)}\in\mathbb{R}^{n\times d}$ are the node embeddings computed after $k$ steps. $\tilde{A}=A+I$. $\tilde{D}=\sum_j\tilde{A}_{ij}$. $W^{(k)}\in\mathbb{R}^{d\times d'}$ is a trainable weighted matrix where $d'$ denotes the output embedding's size. Equation (\ref{eq:gcn}) can be understood as a message passing process. The node embeddings $H^{(k)}$ are the ``messages'' transferred along edges, which are going to be used to generate new node embeddings in next round. A total number of $K$ convolutional layers are stacked to learn node representations and the output matrix $Z=H^K$ can be viewed as the final node representations learned by the GCN model.

In multi-layer GNN, suppose there are totally $L$ layers. At each layer $l$, with the input node feature matrix $X^{l}\in\mathbb{R}^{n^l\times d^l}$ and the adjacency matrix $A^{l}\in\mathbb{R}^{n^l\times n^l}$ generated from previous layer, we learn an embedding matrix $Z^{(l)}\in\mathbb{R}^{n^l\times d^{l+1}}$. Here, we use $d^{l+1}$ to denote the output embeddings' dimension since it will determine the input node embeddings $X^{(l+1)}$ at next layer $l+1$ that will be introduced later. For simplicity's sake, we will use $Z^{(l)}=GCN_{c}(A^{(l)}, X^{l})$ to denote the GCN process (containing $K$ iterations of message passing). Initially, $A^{(1)}=A\in\mathbb{R}^{n\times n}$ is the original graph's adjacency matrix, and $X^{(0)}=F\in\mathbb{R}^{n\times d}$ is the original graph's node features.

In MxPool, we use multiple GCNs to learn multiple sets of node embeddings. These GCNs can be trained with different sets of hyperparameters $\theta$, such as weight matrix $W$'s dimension $d^{l}_i$. Suppose there are $n_c$ GCNs running concurrently at each layer $l$, we will have $n_c$ sets of node embeddings $\{Z_1^{(l)},Z_2^{(l)},\ldots,Z_{n_c}^{(l)}\}$. Let $\theta_i$ be the hyperparameters set of the $i$th GCN. Then at layer $l$, we have node embeddings $Z_i^{(l)}$ resulted from the $i$th GCN as follows:
\begin{equation}\label{eq:mxgcn}
  Z_i^{(l)}=GCN_{c}(A^{(l)}, X^{l}, \theta_i).
\end{equation}
\Paragraph{Use Graph's A Priori Properties for Merging} We then utilize the input graph's a priori properties to merge these diverse sets of node embeddings. In our implementation, we use the number of nodes, the number of edges, and the average node degree to construct a 3-dimensional graph properties vector. Let $\mathfrak{g}\in\mathbb{R}^f$ denote the a priori graph properties vector of a specific input graph where $f$ is the number of graph properties to be considered. According to $\mathfrak{g}$, we employ a softmax normalization to obtain the attention weight of each GCN:
\begin{equation}\label{eq:softmax}
  \alpha^{\mathfrak{g}}_i=\frac{exp(\mathfrak{g}^T\cdot W^g_i)}{\sum_{j=1}^{n_c}exp(\mathfrak{g}^T\cdot W^g_j)},
\end{equation}
where $W^g\in\mathbb{R}^{f\times n_c}$ is the weight matrix of a shared linear transformation which is applied to every input graph. Then, the multiple sets of node embeddings $\{Z_1^{(l)}, Z_2^{(l)},\ldots, Z_{n_c}^{(l)}\}$ are weighted first and merged into one set of node embeddings $Z^{(l)}$ using a neural network:
\begin{equation}\label{eq:mxgcnmerge}
  Z^{(l)}=f_{c}\Big(\alpha^{\mathfrak{g}}_1\otimes Z_1^{(l)}\oplus \alpha^{\mathfrak{g}}_2\otimes Z_2^{(l)}\oplus\ldots\oplus\alpha^{\mathfrak{g}}_{n_c}\otimes Z_{n_c}^{(l)}\Big),
\end{equation}
where ``$\oplus$'' denotes a row-wise concatenation operation, "$\otimes$" denotes a scalar operation, and $f_c()$ is a single-layer MLP neural network. One hyperparameter of $f_c()$ is the output embeddings' dimension $d^{l+1}$, i.e., $Z^{(l)}\in\mathbb{R}^{n^l\times d^{l+1}}$, which is set by averaging the dimensions of the multiple weight matrices $\{W_1,W_2,\ldots,W_{n_c}\}$. Suppose $W_i\in\mathbb{R}^{n^l\times d_i^l}$, we can set $d^{l+1}=\sum_{i=1}^{n_c}d_i^l/n_c$.

\subsection{Multiplex Pooling}

We follow DiffPool \citep{diffpool} to construct our multiplex pooling layer. We learn to assign nodes to clusters at each layer $l$ using the node embeddings and adjacency matrix generated from previous layer. Specifically, at each layer $l$, we learn $n_p$ cluster assignment matrices $\{S_1,S_2,\ldots,S_{n_p}\}$, and each cluster assignment matrix $S_i$ is generated as follows:
\begin{equation}\label{eq:assign}
  S_i^{(l)}=softmax\Big(GCN_{p}(A^{(l)},X^{(l)},\mu_i)\Big).
\end{equation}
It is noticeable that $GCN_{p}$ is a GCN which is different from the $GCN_c$ used in the convolution layer, though these two GNNs consume the same input data. Each row of $S_i^{(l)}$ corresponds to one of the $n^l$ nodes at layer $l$, and each column of $S_i^{(l)}$ corresponds to one of the $c_i^l$ clusters, so that we have $S_i^{(l)}\in\mathbb{R}^{n^l\times c_i^l}$. $\mu_i$ denotes the hyperparameters set of the $i$th GCN. One important hyperparameter could be the coarsening ratio that determines the number of clusters to be assigned. Different pooling networks can have different number of clusters.

These generated assignment matrices $\{S_1^{(l)}, S_2^{(l)},\ldots, S_{n_p}^{(l)}\}$ are merged into a single assignment matrix $S^{(l)}$ in a similar way to that in the convolution operation:
\begin{equation}\label{eq:mxpoolmerge}
  S^{(l)}=f_{p}\Big(\beta^{\mathfrak{g}}_1\otimes S_1^{(l)}\oplus\beta^{\mathfrak{g}}_2\otimes S_2^{(l)}\oplus\ldots\oplus\beta^{\mathfrak{g}}_{n_p}\otimes S_{n_p}^{(l)}\Big),
\end{equation}
where $f_p()$ is a single-layer MLP neural network. Given the number of nodes at the next layer $l+1$, $n^{l+1}$, $f_p()$ should be configured to output an assignment matrix with $n^{l+1}$ columns, i.e., $S^{(l)}\in\mathbb{R}^{n^l\times n^{l+1}}$.

The merged node embeddings $Z^{(l)}$ as shown in Equation (\ref{eq:mxgcnmerge}) and the merged assignment matrix $S^{(l)}$ as shown in Equation (\ref{eq:mxpoolmerge}) are used to generate embeddings for each of the $n^{l+1}$ clusters \citep{diffpool}. The adjacency matrix $A^{(l)}$ and the merged assignment matrix $S^{(l)}$ are also used to generate a coarsened adjacency matrix denoting the edge weights between each pair of cluster:
\begin{equation}\label{eq:finalpool}
\begin{aligned}
  X^{(l+1)}&={S^{(l)}}^TZ^{(l)},\\
  A^{(l+1)}&={S^{(l)}}^TA^{(l)}S^{(l)},
\end{aligned}
\end{equation}
Here, since $S^{(l)}\in\mathbb{R}^{n^l\times n^{l+1}}$ and $Z^{(l)}\in\mathbb{R}^{n^l\times d^{l+1}}$, we have the cluster embeddings $X^{(l+1)}\in\mathbb{R}^{n^{l+1}\times d^{l+1}}$. Similarly, we have the coarsened adjacency matrix $A^{(l+1)}\in\mathbb{R}^{n^{l+1}\times n^{l+1}}$

Note that, the coarsened graph is a fully connected weighted graph, so that the coarsened adjacency matrix $A^{(l+1)}$ is a real matrix and each entry in $A^{(l+1)}$ denotes the edge weight between two clusters. The cluster embeddings $X^{(l+1)}$ and the coarsened adjacency matrix $A^{(l+1)}$ will then be used as input to the next layer, where one cluster at layer $l$ corresponds to one node at layer $l+1$.

\subsection{Computational Complexity Analysis}

Finally we discuss the time complexity of MxPool. Suppose there is totally $L$ layers. At each layer $l$, there are $n_c$ GCNs. The $i$th GCN has $O(m^ld^l)$ time for message passing and $O(n^ld^ld^{l}_i)$ time for matrix multiplication (linear transformation), where $m^l, n^l, d^l$ are respectively the number of edges, the number of nodes, the node embedding size for layer $i$'s input graph and $d^l_i$ is the $i$th GCN's output node embedding size. For simplicity of analysis we assume only one convolution layer exists in each GCN. The total time for $n_c$ GCNs is $O\big(\sum_{i=1}^{n_c}(m^ld^l+n^ld^ld^{l+1}_i)\big)$. The time for learning the attention weights for different GCNs is $O(fn_c)$ where $f$ is the number of considered graph properties. The time for merging $n_c$ GCNs' results is $O\big(n^l\sum_{i=1}^{n_c}d^{l}_i(\sum_{i=1}^{n_c}d^{l}_i/n_c)\big)$ where $\sum_{i=1}^{n_c}d^{l}_i$ is the length of the concatenation of node embeddings. Hence, the total time for multiplex convolution is $T_{Gi}=O\big(\sum_{i=1}^{n_c}(m^ld^l+n^ld^ld^{l+1}_i)+fn_c+n^l(\sum_{i=1}^{n_c}d^{l}_i)^2/n_c\big)$.

The multiplex pooling step has the same GCN process and attention weights assignment process. In addition, the time for generating the $n_p$ assignment matrices is $O\big(n^ld^l\sum_{i=1}^{n_p}c^l_i\big)$ where $c^{l}_i$ is the number of output clusters for the $i$th pooling network, and the time for merging $n_p$ assignment matrices is $O\big(n^l\sum_{i=1}^{n_p}c^l_i(\sum_{i=1}^{n_p}c^l_i/n_p)\big)$ where $\sum_{i=1}^{n_p}c^l_i/n_p=n^{l+1}$. Hence, the time for multiplex pooling at each layer $l$ is $T_{Pi}=O\big(\sum_{i=1}^{n_p}(m^ld^l+n^ld^lc^l_i)+fn_p+n^ld^l\sum_{i=1}^{n_p}c^l_i+n^l(\sum_{i=1}^{n_c}c^{l}_i)n^{l+1}\big)$. The time for generating next layer's node embeddings and adjacency matrix is $T_{Ai}=O\big(n^ln^{l+1}d^{l+1}+(n^ln^{l+1})^2\big)$.

Therefore, the total time is $\sum_{l=1}^L(T_{Gi}+T_{Pi}+T_{Ai})$ where $n^1$ and $m^1$ are the number of nodes and edges of the input graph. $m^1=(n^l)^2$ if $l>1$ since the coarsened graphs are fully connected graphs. MxPool introduces additional convolution networks and pooling networks which brings additional cost. But the main computational cost results from $T_{Ai}$ for generating next layer coarsened graph, especially for large graphs where $n^l$ is large.

\section{Experiments}

In this section, we compare MxPool with the state-of-the-art graph representation learning methods in the context of graph classification task.

\newcommand{\tabincell}[2]{\begin{tabular}{@{}#1@{}}#2\end{tabular}}

\begin{table*}[t]
    %\vspace{-0.1in}
	\caption{Statistics of data sets.}
	\vspace{0.05in}
	\label{tab:data}
	\centering
	{\begin{tabular}{l c c c c c}	
    \toprule[1pt]
		\textbf{dataset} &
		\textbf{graphs} &
		\textbf{classes} &
        \textbf{[min,max] nodes}&
		\textbf{[min,max] edges}&
		\textbf{[min,max] avg-deg} \\
		\toprule[0.5pt]
		{D\&D} & 1178 & 2 & [30, 5748] & [63, 14267] & [7.22, 17.87] \\
		{ENZYMES} & 600 & 6 & [2, 125] & [1, 149] & [2.00, 10.46]\\
		{PROTEINS} & 1113 & 2 & [4, 620] & [5, 1049] & [3.43, 10.14] \\
		{NCI109} & 4127 & 2 & [4, 111] & [3, 119] & [2.50, 5.54] \\	
        {COLLAB} & 5000 & 3 & [60, 492] & [60, 40120] & [13.94, 952.02] \\	
        {RDT-M12K} & 11929 & 11 & [2, 3782] & [1, 5171] & [4.00, 26.37] \\	
    \toprule[1pt]
	\end{tabular}
	}
	\vspace{-0.1in}
\end{table*}

\subsection{Experimental Settings}

\Paragraph{Datasets} In our experiments, we use four bioinformatics
protein datasets: D\&D \citep{ddgraph}, ENZYMES \citep{proteingraph}, PROTEINS \citep{proteingraph}, NCI109 \citep{ncigraph}, and two social network datasets: COLLAB \citep{NIPS2015_5880} and RDT-M12K \cite{rdtm12k}. Each of these datasets include hundreds to thousands graphs. The details of these datasets are provided in Table \ref{tab:data}. The graphs exhibit great diversity on graph sizes and complexity. Table \ref{tab:data} also lists the min/max number of nodes/number of edges/average node degree of each dataset of graphs. The nodes in bioinformatics graphs have categorical features. As regards social graphs, whose nodes do not have features, we use an uninformative feature vector for all nodes. This helps compare all models with the same input representations.

		%{D\&D} & 1178 & 2 & 5748/30/284.32/272 & 14267/63/715.66/693.91 & 17.87/7.22/9.96/1.17 \\
		%{ENZYMES} & 600 & 6 & 125/2/32.63/221.02 & 149/1/62.14/25.50 & 10.46/2/7.73/0.94\\
		%{PROTEINS} & 1113 & 2 & 620/4/39.05/14.96 & 1049/5/72.82/84.60 & 10.14/3.43/7.47/0.85 \\
		%{NCI109} & 4127 & 2 & 111/4/29.68/13.55 & 119/3/32.13/14.96 & 5.54/2.5/4.33/0.21 \\	
        %{COLLAB} & 5000 & 3 & 492/60/74.49/62.30 & 40120/60/2457.78/6439.05 & 952.02/13.94/149.51/175.86 \\	
        %{RDT-M12K} & 11929 & 11 & 3782/2/391.43/428.67 & 5171/1/456.89/518.17 & 952.02/13.94/149.51/175.86 \\

\Paragraph{Baselines} We consider the following state-of-the-art methods for graph classification task as baselines: GraphSAGE \citep{graphsage2017} is a graph convolution framework proposed for semi-supervised node classification. GraphSAGE with global mean-pooling on the learned node representations can realize graph representation learning. SortPool \citep{sortpool2018} is a global pooling method which uses sorting for pooling. It is built upon the GCN layer, where the features of nodes are sorted before feeding them into traditional 1D convolutional and dense layers. gPool \citep{graphunet} achieves pooling operation by adaptively selecting top-k nodes to form a smaller graph based on their scalar projection values on a trainable projection vector. SAGPool \citep{sagpool} is a Self-Attention Graph Pooling method for GNNs in the context of hierarchical graph pooling. The self-attention mechanism is exploited to distinguish between the nodes that should be dropped and the nodes that should be retained. GIN \cite{gin2019} abbr. Graph Isomorphism Network, is shown to be more powerful than traditional GNNs and is as powerful as the Weisfeiler-Lehman graph isomorphism test. DiffPool \citep{diffpool} is the first end-to-end trainable graph pooling method that learns hierarchical representations of graphs.

\Paragraphnopunc{Experimental Setup} In order to remove unwanted bias towards the training data, we use 10-fold cross validation for all the baselines and our approach. Since DiffPool is the key components in our MxPool approach, we use the same hyperparameter as in our MxPool approach. Regarding the hyperparameters of GraphSAGE, SortPool, gPool, SAGPool, and GIN, we follow the same experimental setups described in their original papers. In addition, we adopt the widely used evaluation metric, i.e., accuracy, for graph classification to evaluate the performance. The final test fold score is obtained as the mean of three runs with unfavorable random weight initializations. All models are trained using one NVIDIA GeForce RTX 2080 Ti GPU.

\Paragraph{MxPool Configurations} We implement MxPool using Pytorch. The convolution GNN model used is the original Graph Convolution Network model \cite{nodeclassification}. The pooling GNN model used is the DiffPool model. The model configurations for convolution GNN and pooling GNN are the same as DiffPool. Besides, our MxPool comprises multiple graph convolution networks and multiple graph pooling networks with different sets of hyperparameters to learn graph features from different perspectives. We concurrently run 3 graph convolution networks and also concurrently run 3 graph pooling networks. The a priori graph properties vector used in our model contains 3 graph properties including the number of nodes, the number of edges, and the average node degree. These a priori graph properties for each graph are prepared before the training starts. We train our networks using Adam optimizer with a learning rate of 0.001.

\subsection{Accuracy Results on Graph Classification}

\begin{table*}[t]
\vspace{-0.1in}
	\caption{Average accuracy on graph classification.}
	\vspace{0.05in}
	\label{tab:perf}
	\centering
	{\begin{tabular}{l c c c c c c}
		\toprule[1pt]
		{\textbf{Baselines}} &
		{\textbf{D\&D}} &
		{\textbf{ENZYMES}} &
		{\textbf{PROTEINS}}&
		{\textbf{NCI109}} &
        {\textbf{COLLAB}} &
        {\textbf{RDT-M12K}} \\
		\toprule[0.5pt]
		{BaseLine \citep{faircompare}} & 78.07 & 61.72 & 75.16 & 66.95 & 55.65 &  23.58\\
		{GraphSAGE \citep{graphsage2017}} & 72.36 & 33.25 & 70.48 & 76.50 & 68.25 &  42.20\\
		{SortPool \citep{sortpool2018}} & 78.32 & 31.29 & 73.54 & 70.80 & 73.76 & 31.44\\
		{gPool \citep{graphunet}} & 75.01 & 48.33 & 71.63 & 74.52 & 71.12 & OOR \\
		{SAGPool \citep{sagpool}} & 76.94 & 43.99 & 72.91 & 72.51 & 79.27 & 43.25\\
        {GIN \citep{gin2019}} & 75.57 & 48.32 & 71.65 & 75.44 & \textbf{79.48} & 47.22\\
        {DiffPool \citep{diffpool}} & 80.01 & 62.17 & 75.96 & 80.10 & 71.78 & 47.05\\
        {MxPool (Ours)} & \textbf{81.13} & \textbf{69.53} & \textbf{78.40} & \textbf{83.05} & 77.20 & \textbf{47.52} \\
		\toprule[1pt]
	\end{tabular}
	}
	\vspace{-0.1in}
\end{table*}

We evaluate our proposed MxPool on six benchmark datasets and compare with several state-of-the-art baselines. The baseline method proposed in \citep{faircompare} is structure-agnostic and only exploits node features. The accuracy results are reported in Table \ref{tab:perf} where the best results are shown in bold. For gPool, SAGPool, and GIN, we use their PyTorch Geometric implementations to test graph classification accuracy. Regarding the SAGPool baseline \citep{sagpool}, we meet OutOfResource error when processing the RDT-M12K dataset, and we denote this case by `OOR'.

%For all the baselines, we use 10-fold cross validation numbers reported by the original authors if we can obtain the numbers close to the reported ones. Regarding the gPool baseline \citep{graphunet}, we cannot obtain the necessary published numbers, so we use the numbers on the ENZYMES dataset and PROTEINS dataset reported by the third-party\footnote{https://github.com/bknyaz/graph\_nn} and the number on D\&D dataset reported by \citet{sagpool}. For the other datasets, we use the numbers tested by ourselves. Regarding the SAGPool baseline \citep{sagpool}, we meet RuntimeError when processing the ENZYMES dataset and COLLAB dataset, and we denote this case by `-'.

From the table, we observe that our proposed MxPool approach obtains the best performance on 5 out of 6 benchmark datasets. GIN and SAGPool are slightly better than MxPool on the COLLAB dataset. But MxPool performs better than GIN and SAGPool on the other datasets, e.g., MxPool improves upon GIN and SAGPool by an average of 6.53\% and 7.99\%, respectively. The DiffPool model is the most similar one to our MxPool. We extend the DiffPool model by employing the multiplex structure. Our model outperforms DiffPool by an average of 3.29\%, which can be attributed to the effect of the multiplex structure. Specially, MxPool shows significant performance improvement over DiffPool on multi-class classification. For example, MxPool outperforms DiffPool on the ENZYMES dataset (with 6 classes) by 7.36\% and on the COLLAB dataset (with 11 classes) by 5.42\%, while only outperforms DiffPool on the other datasets by an average of 1.75\%.

%shows remarkable performance improvement over the other state-of-the-art baselines on the first four datasets. Especially on the D\&D dataset, MxPool improves the performance of the second-best approach DiffPool over 11\%. However, DiffPool performs slightly better than MxPool on the COLLAB dataset. We tested DiffPool by ourselves and could only obtain 71.78\% accuracy, which is lower than the number 75.48\% reported in the original paper. This may be due to the unoptimized parameter settings, but we still use the reported number 75.48\% for the DiffPool baseline. In addition, the DiffPool model is the most similar one to our MxPool. We extend DiffPool by employing the multiplex structure. We can see that MxPool improves upon the base DiffPool by an average of 3.62\%, which can reflect the effectiveness of the multiplex structure.

%There could exist bugs for some baselines making it is hard to obtain correct results for some datasets, where we will use ``-'' to label these phenomena.

\subsection{Effects of Multiplex Convolution/Pooling}

Our motivation for this work is to utilize multiplex hierarchical structure to deal with the diversity and complexity challenges in graph representation learning. Multiple graph convolutional networks with different sets of hyperparameters are used to learn node representations. The node embedding size, as a hyperparameter in GCN, plays an important role in determining the quality of node representation. We vary the node embedding sizes in different GCN networks. On the other hand, multiple graph pooling networks (i.e., DiffPool) with different sets of hyperparameters are used to coarsen graphs. The compression ratio, as a hyperparameter in DiffPool, plays an important role in determining the quality of graph representation. We vary the compression ratios in different DiffPool networks.

\begin{table*}[t]
%\vspace{-0.1in}
	\caption{Effects of multiplex convolution/pooling.}
	\vspace{0.05in}
	\label{tab:effect}
	\centering
	{\begin{tabular}{l c c c c c c}
		\toprule[1pt]
		\multicolumn{2}{c}{\textbf{Variations}} &
		{\textbf{D\&D}} &
		{\textbf{ENZYMES}} &
		{\textbf{PROTEINS}}&
		{\textbf{NCI109}} &
        {\textbf{COLLAB}} \\
        \toprule[0.5pt]
		\multirow{3}*{SCSP} & {$c_1*p_1$} & 80.01 & 61.42 & 75.48 & 80.07 & 71.31\\
		~ & {$c_2*p_1$} & 79.33 & 62.17 & 74.47 & 80.10 & 71.33\\
		~ & {$c_3*p_1$} & 78.75 & 61.44 & 75.14 & 78.49 & 71.14\\
        \toprule[0.5pt]
		MCSP & {$[c_1|c_2|c_3]*p_1$} & 80.47 & 68.22 & 76.30 & 81.89 & 73.43\\
        \toprule[0.5pt]
        \multirow{3}*{SCSP} & {$c_1*p_1$} & 80.01 & 61.42 & 75.48 & 80.07 & 71.31\\
        ~ & {$c_1*p_2$} & 78.64 & 60.75 & 75.96 & 78.47 & 71.78\\
        ~ & {$c_1*p_3$} & 78.79 & 61.58 & 74.52 & 78.95 & 71.02\\
        \toprule[0.5pt]
        SCMP & {$c_1*[p_1|p_2|p_3]$} & 80.06 & 63.02 & 76.01 & 80.30 & 73.05\\
        \toprule[0.5pt]
        MCMP & {$[c_1|c_2|c_3]*[p_1|p_2|p_3]$} & \textbf{81.13} & \textbf{69.53} & \textbf{78.40} & \textbf{83.05} & \textbf{77.20} \\
		\toprule[1pt]
	\end{tabular}
	}
	\vspace{-0.1in}
\end{table*}

In order to verify the effectiveness of multiplex convolution and multiplex pooling, we run our MxPool with single convolution network and single pooling network (SCSP), multiple convolution networks and single pooling network (MCSP), single convolution network and multiple pooling networks (SCMP), and multiple convolution networks and multiple pooling networks (MCMP), respectively. The accuracy results on 5 datasets for graph classification are shown in Table \ref{tab:effect}.
Since the suitable node embedding sizes and compression ratios are not consistent for different datasets, we use $c_1,c_2,c_3$ to denote three different graph convolution parameters (i.e., node embedding size) and $p_1,p_2,p_3$ to denote three different graph pooling parameters (i.e., coarsening ratio). Note that, they are different for different datasets. We have put the detailed parameter settings on our GitHub project page.

From the table, we observe that the multiplex structure significantly improves performance over the singular structure. By fixing the pooling network with $p_1$, multiplexing three convolution networks with hyperparameters $[c_1|c_2|c_3]$ consistently performs better than using single convolution network with either $c_1$, $c_2$, or $c_3$. A similar trend can be observed when multiplexing pooling networks. Anyhow, the best choice is to simultaneously multiplex convolution networks and multiplex pooling networks (i.e., MCMP). Both multiplex convolution and multiplex pooling play important role in improving performance, but the best choice is to use them together.

\subsection{Distribution of Attention Weights}

\begin{figure*}[t]
	\centerline{
	\subfigure[Graphs with various node numbers]{\includegraphics[width=2.3in]{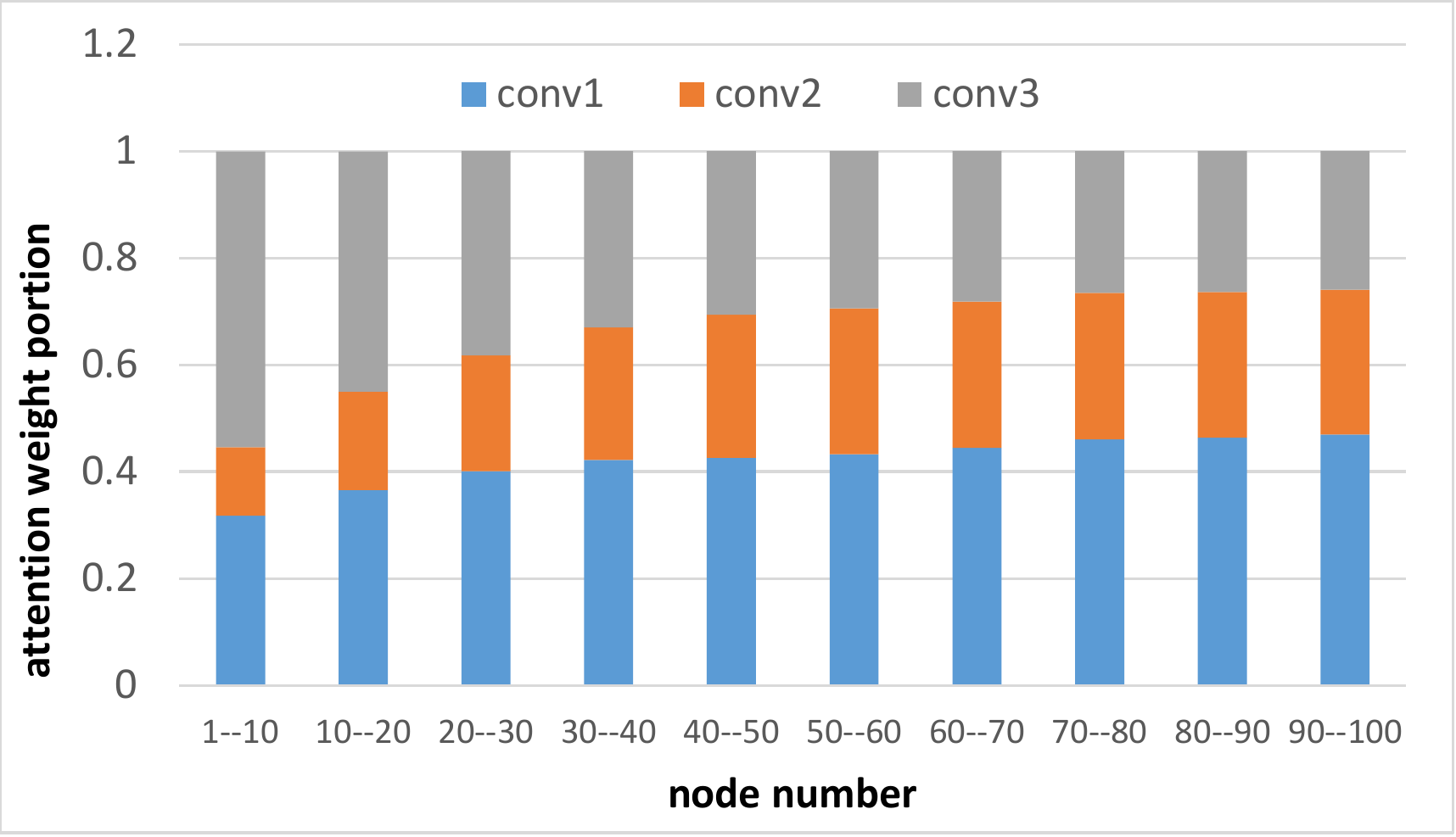}
    \label{fig:convnode}
    \vspace{-0.05in}}
    \subfigure[Graphs with various edge numbers]{\includegraphics[width=2.3in]{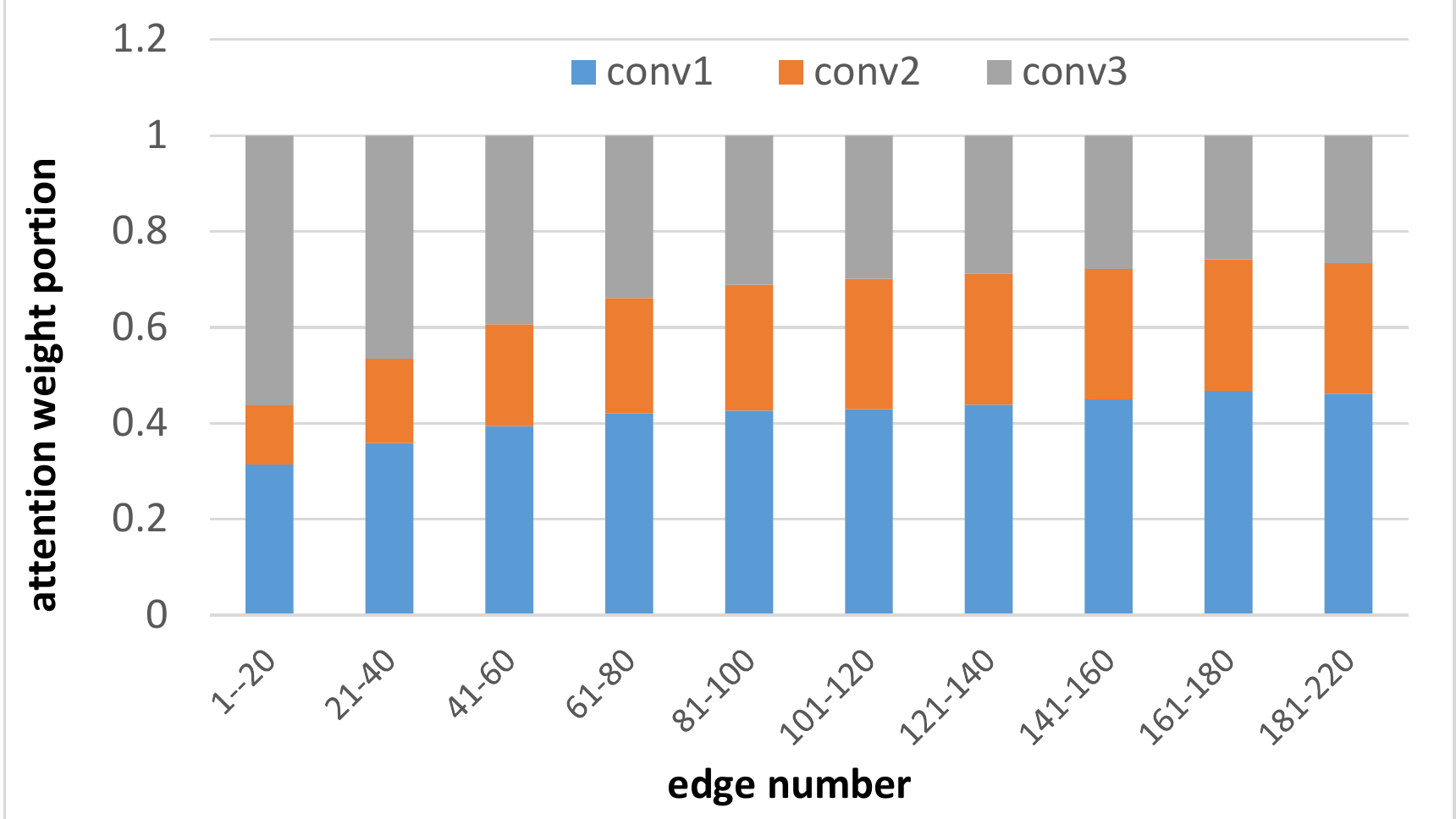}
    \label{fig:convedge}
    \vspace{-0.05in}}
    \subfigure[Graphs with various avg. degrees]{\includegraphics[width=2.3in]{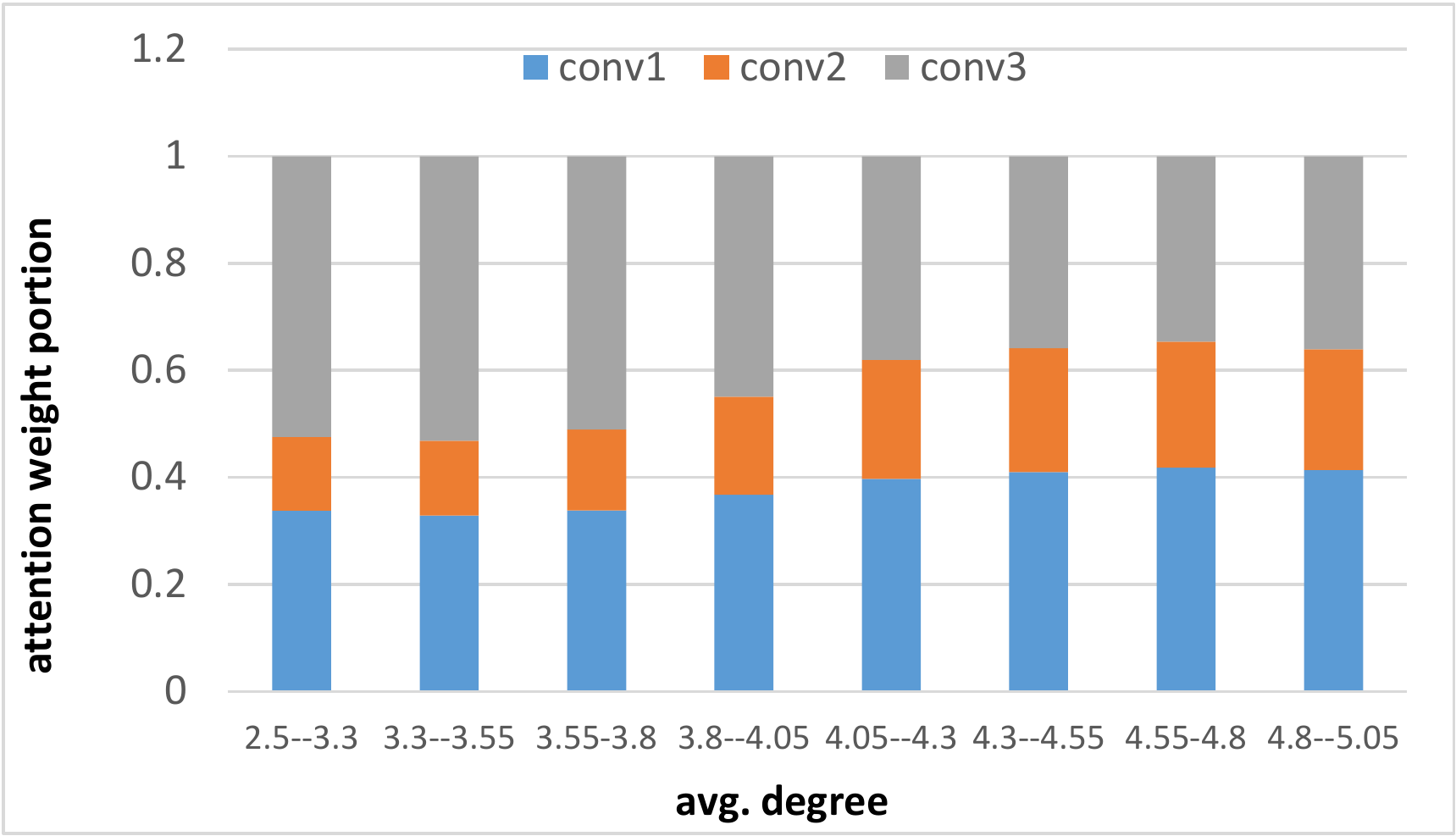}
    \label{fig:convdegree}
    \vspace{-0.05in}}
    }
    %\vspace{-0.1in}
	\caption{Attention weight distribution of convolution networks (NCI109)}
	\label{fig:conv}
%\vspace{-0.1in}
\end{figure*}

\begin{figure*}[t]
	\centerline{
	\subfigure[Graphs with various node numbers]{\includegraphics[width=2.3in]{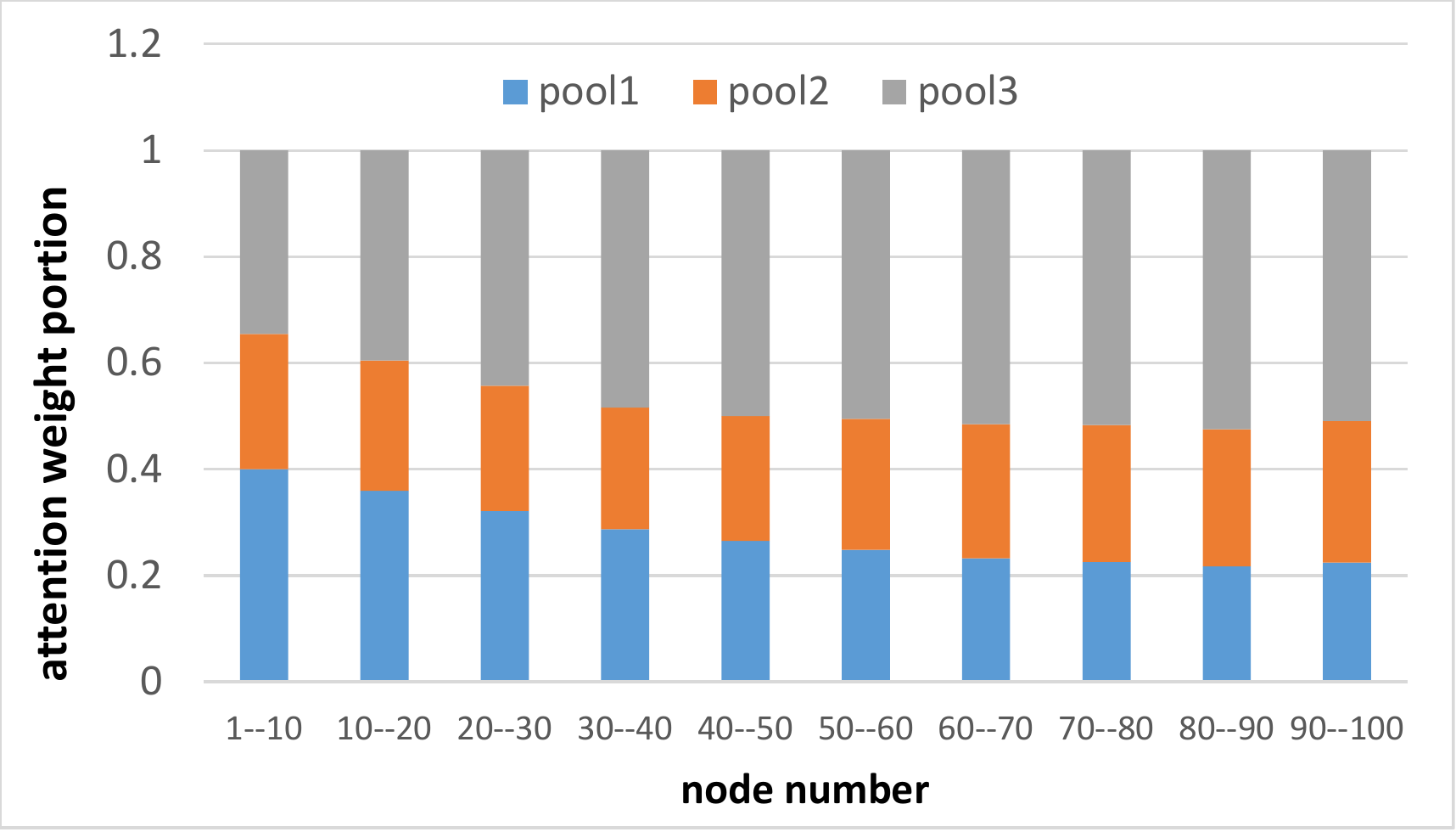}
    \label{fig:poolnode}
    \vspace{-0.05in}}
    \subfigure[Graphs with various edge numbers]{\includegraphics[width=2.3in]{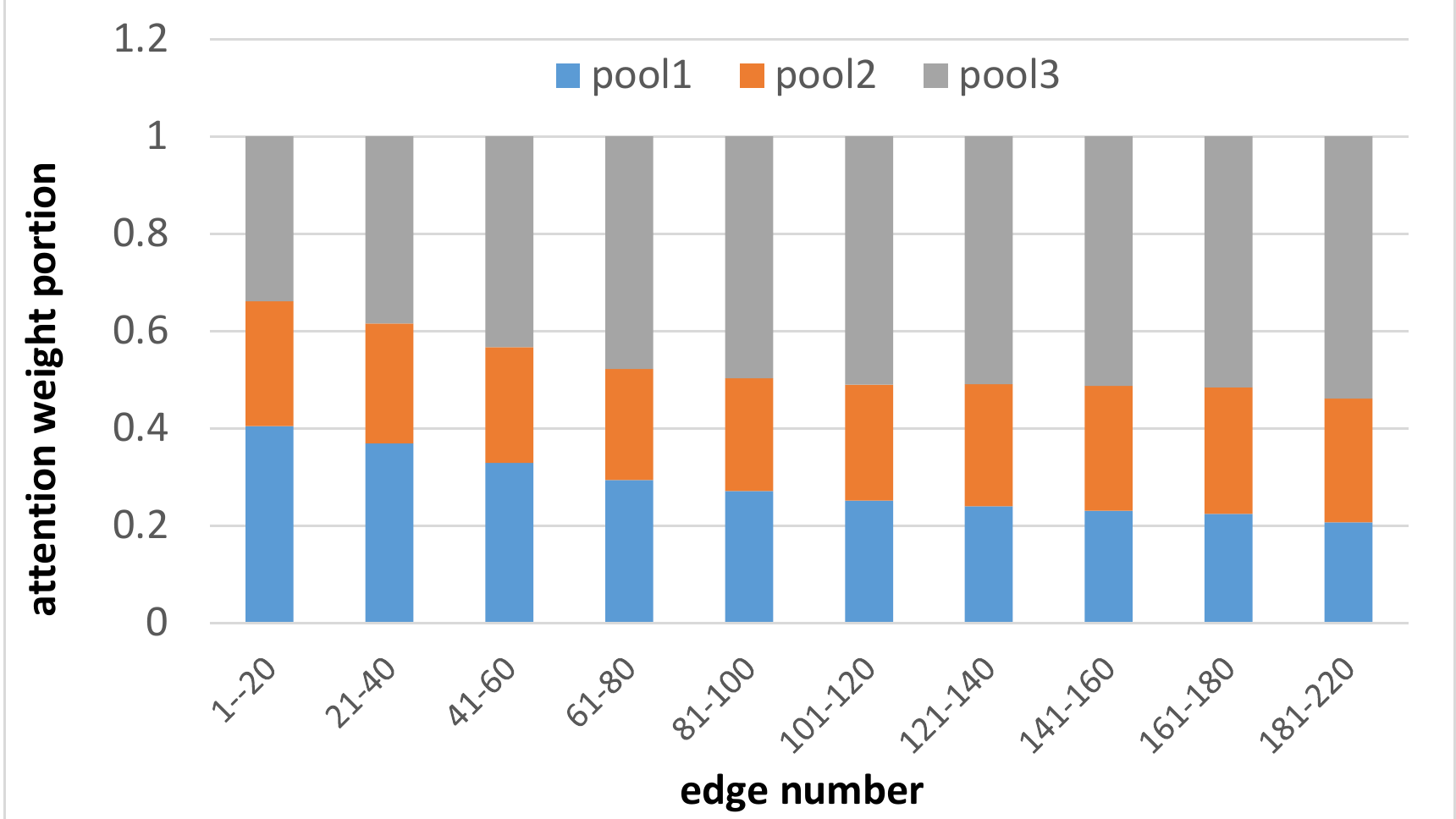}
    \label{fig:pooledge}
    \vspace{-0.05in}}
    \subfigure[Graphs with various avg. degrees]{\includegraphics[width=2.3in]{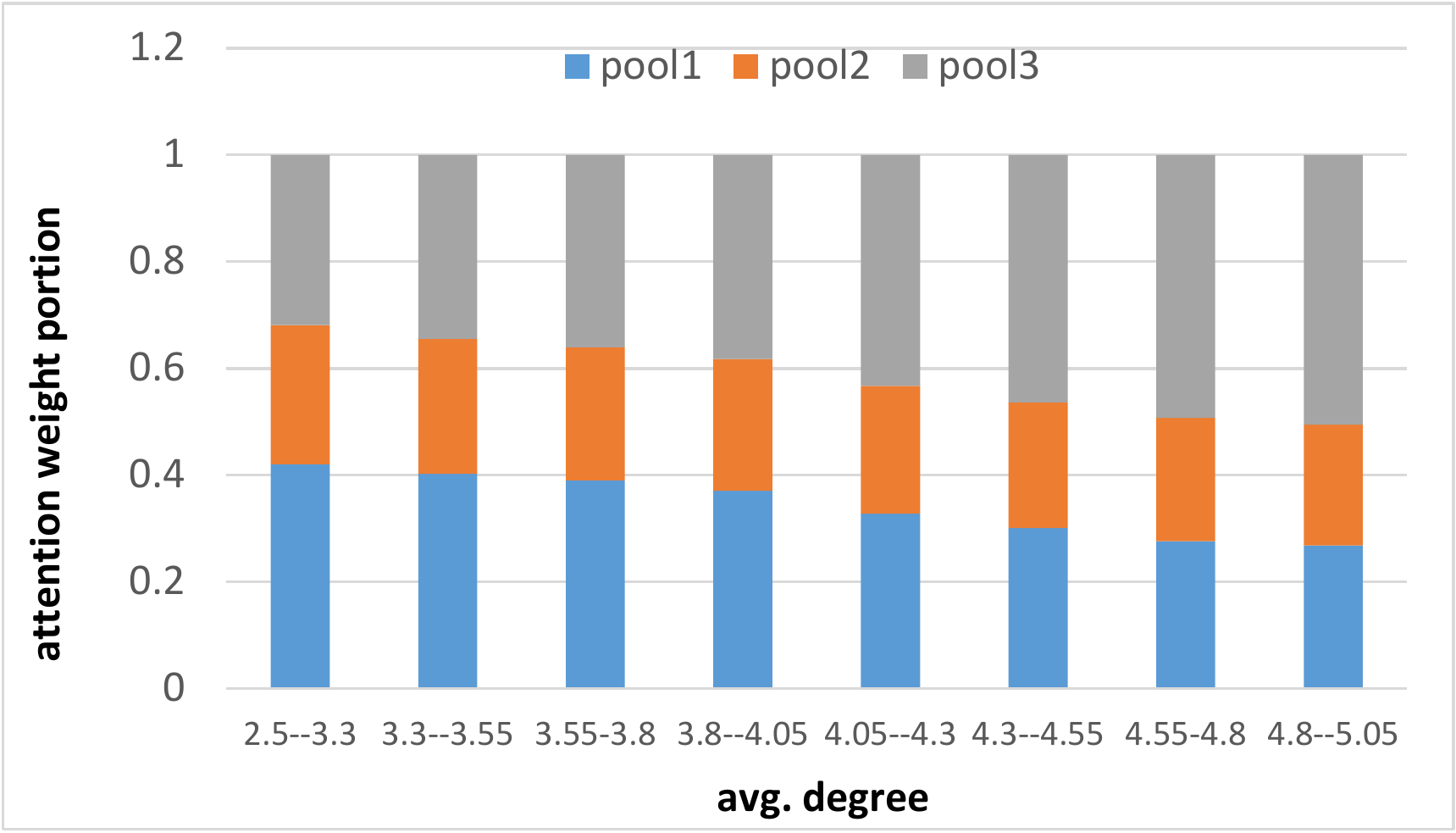}
    \label{fig:pooldegree}
    \vspace{-0.05in}}
    }
    %\vspace{-0.1in}
	\caption{Attention weight distribution of pooling networks (NCI109)}
	\label{fig:pool}
%\vspace{-0.1in}
\end{figure*}

In order to learn graph features from a diverse graphs dataset where graphs exhibit very different properties, MxPool employs multiple convolution networks and multiple pooling networks to learn graph features from different aspects. By exploiting the a prior graph properties (e.g., number of nodes, number of edges, and average degree), MxPool assigns different attention weights to different convolution networks and assigns different attention weights to different pooling networks. In order to verify if the attention weights can be learned through our end-to-end structure, we first run the MxPool training process on the NCI109 dataset and record the learned attention weights after the training process is completed.

The recorded attention weights distribution of multiple convolution networks is shown in Figure \ref{fig:conv}. We group the attention weight results based on graph's node number and show the average attention weight of each group in the figure. That is, we show the attention weights distribution for different size graphs, so that we can see if the weights distribution varies when processing different size graphs. The attention weight portion of each convolution networks is depicted in Figure \ref{fig:convnode}. We observe that graphs with different node numbers indeed obtain different attention weight distributions. It also clearly shows that the attention weight to each convolution network is relevant to the graph size (number of nodes in the graph). Figure \ref{fig:convedge} and Figure \ref{fig:convdegree} also show similar trends when processing graphs with different edge numbers and with different average node degrees.

The recorded attention weights distribution of multiple pooling networks is shown in Figure \ref{fig:pool}. We can also observe the similar trend as the multiple convolution networks as shown in Figure \ref{fig:conv}. These results verify that our MxPool indeed have learned different attention weights for different convolution/pooling networks, which supports our idea of constructing multiplex structures, because the multiplex structure can help learn graph features from different size graphs.

\subsection{Number of Convolution/Pooling Networks}

The number of convolution/pooling networks is a hyperparamter in MxPool. In the previous experiments, we use a fixed number of convolution/pooling GNNs to show the performance. In this experiment, we vary the number of convolution/pooling networks from 1 to 5 and evaluate the performance. The number of convolution networks and the number of pooling networks are the same. The graph classification accuracy results on ENZYMES and PROTEINS datasets are listed in Table \ref{tab:nGNN}. From the table, we can see that the best performance is achieved when the number is set around 3-4. As the number is increased larger, the performance is reduced. This may be because that too many networks with a large amount of parameters result in overfitting problem.

\begin{table}[t]
    %\vspace{-0.1in}
	\caption{Effect of number of convolution/pooling GNNs.}
	\vspace{0.05in}
	\label{tab:nGNN}
	\centering
	{\begin{tabular}{l c c c c c}	
    \toprule[1pt]
		{\textbf{Dataset}} &
		{\textbf{\tabincell{c}{1}}} &
		{\textbf{\tabincell{c}{2}}} &
        {\textbf{\tabincell{c}{3}}}&
		{\textbf{\tabincell{c}{4}}}&
		{\textbf{\tabincell{c}{5}}}
        \\
		\toprule[0.5pt]
		{ENZYMES} & 62.53 & 59.83 & \textbf{69.53} & 65.33 & 61.71\\
		{PROTEINS} & 76.25 & 74.31 & \textbf{78.40} & 77.36 & 76.64\\
    \toprule[1pt]
	\end{tabular}
	}
	%\vspace{-0.1in}
\end{table}

\section{Conclusion}

In this paper, we proposed a simple but effective multiplex GNN architecture MxPool for hierarchical graph representation learning. MxPool comprises multiple graph convolution networks to learn node-level representations and also comprises multiple graph pooling networks to coarsen the graph. The a priori graph properties are employed to assign the attention weight to each convolution/pooling network, so that the diversity challenge of graph representation learning can be well addressed. Our results show that MxPool has gained performance improvement over the state-of-the-art graph representation learning methods. Future work includes designing unpooling layers to form an encoder-decoder learning structure to deal with node classification tasks and link prediction tasks.

\bibliography{gnn}
\bibliographystyle{icml2020}

\end{document}